\documentclass[sigconf, letterpaper]{acmart}
\usepackage{algorithm}
\usepackage{algorithmic}
\usepackage{amsmath}
\usepackage{multicol}
\usepackage{multirow}
\usepackage{colortbl}
\usepackage{tcolorbox}
\usepackage{subfigure}
\usepackage{hyperref}
\usepackage{CJKutf8}
\usepackage{geometry}
\usepackage{xurl} 
\usepackage{hyperref}
\usepackage{balance}
\usepackage{tcolorbox}
\tcbuselibrary{listings,skins,breakable}
\usepackage{listings}

\newtcolorbox{promptbox}[1]{%
enhanced, breakable,
colback=gray!5, colframe=gray!50, boxrule=0.5pt,
sharp corners, drop shadow,
title={#1}, fonttitle=\bfseries,
coltitle=white, colbacktitle=gray!65,
boxed title style={sharp corners},
attach boxed title to top left={yshift=-2mm,xshift=2mm},
listing only,
listing options={basicstyle=\ttfamily\small,breaklines=true}
}

\DeclareMathOperator*{\argmin}{arg\,min}

\AtBeginDocument{%
  }

\copyrightyear{2026}
\acmYear{2026}
\setcopyright{cc}
\setcctype{by-nc-nd}
\acmConference[KDD 2026] {Proceedings of the 32nd ACM SIGKDD Conference on Knowledge Discovery and Data Mining V.2}{August 9--13, 2026}{Jeju Island, Republic of Korea.}
\acmBooktitle{Proceedings of the 32nd ACM SIGKDD Conference on Knowledge Discovery and Data Mining V.2 (KDD 2026), August 9--13, 2026, Jeju Island, Republic of Korea}
\acmISBN{979-8-4007-2259-2/2026/08}
\acmDOI{10.1145/3770855.3818413}
\begin{document}

\title{FOUNDv2: Learning Unified User Quantized Tokenizers for User Representation}




\author{Chuan He}
\authornote{These authors contributed equally to this research.}
\affiliation{%
  \institution{Ant Group}
  \institution{Zhejiang University of Technology}
  \city{Hangzhou}
  \country{China}
}
\email{hechuan@zjut.edu.cn}

\author{Yang Chen}
\authornotemark[1]
\affiliation{%
  \institution{Ant Group}
  \city{Shanghai}
  \country{China}
}
\email{cy462023@antgroup.com}

\author{Bin Dou}
\authornotemark[1]
\affiliation{%
  \institution{Ant Group}
  \city{Hangzhou}
  \country{China}
}
\email{doubin.dou@antgroup.com}

\author{Wuliang Huang}
\affiliation{%
  \institution{Ant Group}
  \city{Shanghai}
  \country{China}
}
\email{huangwuliang.hwl@antgroup.com}





\author{Baokun Wang}
\authornote{Corresponding author}
\affiliation{%
  \institution{Ant Group}
  \city{Hangzhou}
  \country{China}
}
\email{yike.wbk@antgroup.com}

\author{Yongchao Liu}
\affiliation{%
  \institution{Ant Group}
  \city{Hangzhou}
  \country{China}
}
\email{yongchao.ly@antgroup.com}

\author{Xing Fu}
\affiliation{%
  \institution{Ant Group}
  \city{Hangzhou}
  \country{China}
}
\email{zicai.fx@antgroup.com}

\author{Yu Cheng}
\affiliation{%
  \institution{Ant Group}
  \city{Shanghai}
  \country{China}
}
\email{cy122623@antgroup.com}

\author{Chuntao Hong}
\affiliation{%
  \institution{Ant Group}
  \city{Hangzhou}
  \country{China}
}
\email{chuntao.hong@antgroup.com}

\author{Weiqiang Wang}
\affiliation{%
  \institution{Ant Group}
  \city{Hangzhou}
  \country{China}
}
\email{wang.weiqiang@antgroup.com}

\author{Zhongle Xie}
\affiliation{%
  \institution{Zhejiang University}
  \city{Hangzhou}
  \country{China}
}
\email{xiezl@zju.edu.cn}

\author{Jiajun Zheng}
\affiliation{%
  \institution{Ant Group}
  \city{Shanghai}
  \country{China}
}
\email{zjj517361@antgroup.com}

\author{Xin-Wei Yao}
\affiliation{%
  \institution{Zhejiang University of Technology}
  \city{Hangzhou}
  \country{China}
}
\email{xwyao@zjut.edu.cn}

\renewcommand{\shortauthors}{Chuan He et al.}

\begin{abstract}
User representation learning serves as a fundamental pillar for personalized services on large-scale web platforms. Despite its importance, conventional continuous embedding methods face significant challenges, including the lack of a unified paradigm for multi-source data integration, prohibitive storage overhead due to low information density, and the lack of multi-scale modeling granularity. To overcome these limitations, we introduce FOUNDv2, a comprehensive user representation scheme centered on the Unified User Quantized Tokenizer (U\textsuperscript{2}QT) framework. FOUNDv2 transforms heterogeneous user data into a standardized discrete token space through a robust two-stage architecture. Specifically, the framework first extracts compact feature representations and subsequently employs a multi-view RQ-VAE to discretize them into storage-efficient tokens using shared and source-specific codebooks. To empower these representations with predictive intelligence, we further design multi-scale alignment objectives to capture both fine-grained behavioral dependencies and macro-temporal periodicity. Extensive experiments on various benchmarks demonstrate that FOUNDv2 consistently outperforms task-specific baselines while achieving substantial reductions in storage and computational costs. Finally, the large-scale deployment of FOUNDv2 on Alipay validates its practical scalability and efficiency across diverse industrial scenarios.
The main code is available at: \href{https://github.com/chuanhe1999/FOUNDv2}{\url{https://github.com/chuanhe1999/FOUNDv2}}.

\end{abstract}





\maketitle

\section{Introduction}\label{section:intro}
\begin{figure}
    \centering
    \includegraphics[width=1.1\linewidth]{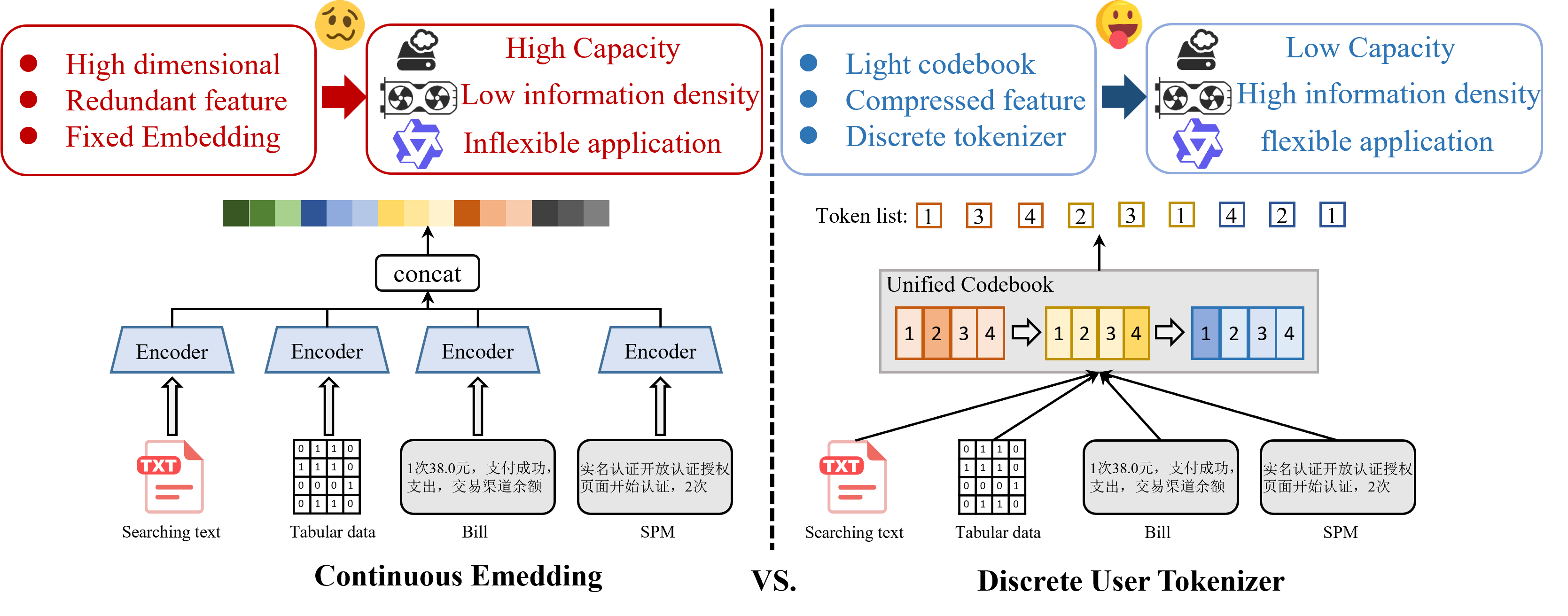}
    \caption{This figure shows the difference between discrete user tokenizer and continuous embedding. Our proposed method benefits from low capacity, high information density and flexible application.}
    \label{fig:advantage}
\end{figure}
In modern hyper-scale web platforms such as Alipay, user modeling serves as the foundational infrastructure for a myriad of downstream applications, ranging from personalized recommendations to real-time fraud detection. These platforms generate a staggering volume of multi-source heterogeneous data, including textual profiles, behavioral sequences, and structured tabular attributes. The ability to distill these diverse inputs into robust, universal user representations is not merely a technical challenge but a strategic necessity for optimizing user experience and platform ecosystem health at scale

To tackle this, the prevailing technical route has shifted from simple statistical features to sophisticated late-fusion paradigms~\cite{usermodelingsurvey24, wu2019context, tang2021disentangleduser}. Early attempts like User2Vec~\cite{hallac2019user2vec} focused on aggregating source-specific embeddings via pooling, while subsequent works like MFN~\cite{bian2021novel} and MSSR~\cite{lin2024multi} introduced complex attention mechanisms to capture inter-source interactions. With the advent of large-scale pre-training, general-purpose frameworks like One4all~\cite{shin2021one4all} and sequence-centric models like MSDP~\cite{fu2023robust} have further pushed the boundaries of representation quality. Most recently, FOUND~\cite{dou2025transferable} reached a new milestone by aligning user behaviors with natural language to facilitate zero-shot transfer. Despite their empirical success, these SOTA methods fundamentally operate on a "separate-encoding-then-fusion" architecture. While this modularity simplifies training, it inherently limits the model's ability to capture deep, cross-modal synergies, as each data source is initially processed in isolation.

We argue that this dominant late-fusion paradigm, including its most advanced variants, has reached a bottleneck characterized by three systemic limitations:

(i) \emph{Lack of unified representation for heterogeneous data}: As illustrated in Figure~\ref{fig:advantage}, conventional methods typically employ specialized encoders for each data source and aggregate them via simple concatenation. This fragmented architecture treats different modalities as isolated entities, failing to capture the \textbf{intrinsic synergies} and complex dependencies across diverse data sources. Consequently, a unified framework is required to seamlessly integrate heterogeneous inputs into a cohesive representation space, effectively balancing source-specific nuances with holistic cross-source interactions.

(ii) \emph{High data capacity yet low information density}: We empirically observe that increasing the temporal breadth of behavioral data within a fixed-size embedding leads to significant performance gains, suggesting that denser representations are inherently more expressive. This aligns with the \textbf{densing laws} of LLMs~\cite{densinglaw}, which emphasize enhancing information density under constrained budgets. However, existing paradigms rely on high-dimensional but sparse continuous vectors, creating a bottleneck that prevents platforms from effectively scaling to long-term user history.

(iii) \emph{Lack of multi-scale modeling granularity}: Traditional continuous-vector paradigms~\cite{yuan2021one, kim2023task} treat user representation as a monolithic, static embedding. Such representation lacks an internal discrete structure, making it mathematically challenging to capture user dynamics across varying temporal scales. Without a unified tokenizer for heterogeneous behaviors, these models cannot effectively bridge the gap between micro-level event transitions and macro-level interest evolution. Consequently, they fail to leverage the hierarchical nature of human behavior, which is essential for understanding complex user trajectories in large-scale platforms.

To address these challenges, we propose \textbf{FOUNDv2}, a unified user representation scheme centered on the \textbf{U$^2$QT} (Unified User Quantized Tokenizers) framework. FOUNDv2 transforms heterogeneous multi-source data into a standardized discrete vocabulary through a two-stage architecture. Specifically, the first stage projects diverse features, including textual descriptions, sequential behaviors, and tabular attributes, into a cohesive semantic space using a pretrained embedding model \cite{qwen3emb}. This process effectively resolves the fragmentation of disparate data sources while preserving rich semantic nuances. The second stage introduces \textbf{MRQ-VAE} (Multi-View Residual Quantized Variational AutoEncoder), which discretizes continuous embeddings into tokens managed by hierarchical shared and source-specific codebooks. This hierarchical design allows the model to simultaneously capture cross-domain commonalities and source-unique behavioral patterns.

This quantization paradigm achieves an \textbf{30$\times$ reduction in memory footprint} and \textbf{3.5$\times$ faster training speeds} compared to the continuous-embedding-based FOUND \cite{dou2025transferable}. By eliminating information redundancy, FOUNDv2 extends the manageable historical window from 60 to \textbf{180 days} under identical GPU memory constraints, representing a \textbf{200\% increase in temporal context} that significantly enhances the modeling of long-term user dynamics. Beyond basic quantization, we further empower U\textsuperscript{2}QT through multi-scale pretext tasks. Specifically, we implement token-level future behavioral alignment to strengthen dependencies within single behavioral events, and window-level alignment to capture the long-range cyclicity of user history. Furthermore, drawing inspiration from CLIP-style contrastive learning, we perform semantic alignment to ensure that the learned user tokens remain predictive of future behavioral patterns at a high-level conceptual space.

To assess the effectiveness and cross-task adaptability of our approach, we introduce a multi-task e-commerce benchmark encompassing user profile, future behavior prediction, and downstream recommendation tasks. Extensive experiments demonstrate superior adaptability and scalability across diverse tasks compared to state-of-the-art baselines. In summary, our contributions are listed as follows,
\begin{itemize}
    \item We propose \textbf{FOUNDv2}, a comprehensive user representation scheme based on our \textbf{MRQ-VAE} technique. To the best of our knowledge, this represents the first framework to transform multi-source heterogeneous user data into a standardized and discrete token space.

    \item Compared to SOTA models, FOUNDv2 achieves an 30$\times$ memory reduction and 3.5$\times$ training speedup, while its 200\% temporal context extension facilitates capturing long-term user dynamics more effectively.
    
    \item We design fine-grained pretext task including token-, window-, and semantic-level future behavior alignment to endow user tokenizer with multifaceted temporal intelligence.

    \item Our framework ensures robust cross-task generalization, consistently outperforming state-of-the-art baselines on real-world multi-task e-commerce benchmark and industrial environment.

\end{itemize}

\section{Related Work}
\subsection{Quantitative Language}
Quantitative language, also referred to as semantic IDs~\cite{pan2024auto,tokenizerentity}, has gained significant traction in VLLMs~\cite{li2025surveystateartlarge,atoken,semhitok, unicode2, unitok, janus-pro} and generative recommendation~\cite{li2024surveygenerativesearchrecommendation, onerec, onesearch,VQ-Rec, tiger, reg4rec, LC-Rec}. 
In recommendation systems, models like MMGRec~\cite{MMGRec} and MQL4GRec~\cite{MQL4GRec} utilize quantization (e.g., RQ-VAE) to unify multimodal features and facilitate knowledge transfer. 
Parallelly, visual tokenization research focuses on balancing reconstruction and semantics; for instance, SEED~\cite{ge2023planting} emphasizes LLM-compatible 1D causal tokens, while UniTok~\cite{unitok} and SemHiTok~\cite{semhitok} leverage multi-codebook or hierarchical structures to decouple high-level semantics from low-level pixels. 
These discrete paradigms effectively mitigate objective conflicts, enabling unified modeling within generative frameworks.
\subsection{Universal User Representation}
User modeling aims to construct accurate user profiles by extracting latent features from diverse behavioral data \cite{usermodelingsurvey24}. Early research in universal user representation learning utilized specialized architectures, such as the attention-based RNNs in DUPN \cite{ni2018perceive} for multitask learning and parameter-efficient transfer frameworks in PeterRec \cite{yuan2020parameter} for cold-start scenarios. Subsequent works, such as SUMN \cite{gu2021exploiting}, introduced self-supervised objectives and multi-hop aggregation to better integrate heterogeneous behavioral signals. To handle evolving task requirements, continual learning paradigms have been developed to mitigate catastrophic forgetting. For instance, Conure \cite{yuan2021one} employs iterative weight pruning to retain cross-task knowledge, while TERACON \cite{kim2023task} utilizes task-specific soft masks and pseudo-labeling to capture task relationships during sequential learning. 

Recently, the linguistic capabilities of Large Language Models (LLMs) have shifted the paradigm toward modeling in the language space. NoteLLM and its successor \cite{notellm, notellm2} focus on item-centric embeddings by fine-tuning LLMs on correlated item pairs through contrastive objectives. Expanding this to user modeling, HLLM \cite{hllm} proposes a joint framework employing stacked LLMs to interpret behavioral sequences as linguistic inputs. These advancements highlight the potential of unified, scalable architectures for universal user representation.

\section{Preliminaries}
\subsection{Concepts}
In the real-world application, extensive non-sensitive user information and interactions are accessible. For instance, user behavioral sequence data is denoted as $\mathcal{V}=\{(B_n, S_n, M_n, A_n)\}_{n=1}^{N}$, comprising PayBill $B$, Super Position Model $S$, MiniProgram $M$, and App $A$. Here, $N$ denotes the number of users and $n\in[1,2,...,N]$. $B_n\in \mathcal{T}^{L_n^{(B)}}$, $S_n\in \mathcal{T}^{L_n^{(S)}}$, $M_n\in \mathcal{T}^{L_n^{(M)}}$ and $A_n\in \mathcal{T}^{L_n^{(A)}}$ represent the raw text for PayBill information, Super Position Model details, MiniProgram, and App descriptions, respectively, where $\mathcal{T}$ is the token dictionary, and $L_n^{(M)}$ indicates the sequence length. Furthermore, tabular data is denoted as $T\in \mathbb{R}^{N\times F\times D}$, where $F$ represents the number of features and $D$ denotes the dimensionality of each feature. Search text data is represented as $R_n\in\mathcal{T}^{L_n^{(R)}}$. Each user's information can be represented as $X_n = (V_n, T_n, R_n)\in \mathcal{X}$, where $V_n \in \mathcal{V}$ and $\mathcal{X}$ is the joint space of all possible user data. This work focuses on learning unified and robust user representations $U$ from these multi-source, multi-source data to support diverse downstream tasks.

\subsection{Problem Definition}
The task of unified user representation involves the integration of extensive multi-source behavioral data into a shared latent space:
\begin{equation}
g_{\theta^{*}}=\argmin_{\theta}\mathop{\mathbb{E}}\limits_{X_{n}^{s}\in \mathcal{X}^{s}}\mathcal{L}_{\text{pre}}(g_{\theta};X_{n}^{s}),
\end{equation}
where $X_{n}^{s}$ is the multi-source user data used for pretraining, $g_{\theta^{*}}$ represents the pretrained unified user representation, and $\mathcal{L}_{\text{pre}}$ denotes the proposed pretraining task, which will be illustrated in Section~\ref{method}.

With the pretrained unified user representations $U=g_{\theta^{*}}(X)$, we can utilize them for various downstream tasks:
\begin{equation}
    \hat{y}_{n}^i=\mathcal{P}_{\omega_{i}}(U_n)
\end{equation}
where $\hat{y}_{n}^i$ is the predicted label of $n$-th user in the downstream task $i\in[1,2,...,I]$, $\omega_i$ is the model parameters for specific tasks. In summary, this is a two-stage framework in which user representations are generated first and fed into a specific model for downstream task.

\begin{figure*}
    \centering
    \includegraphics[width=1.0\linewidth]{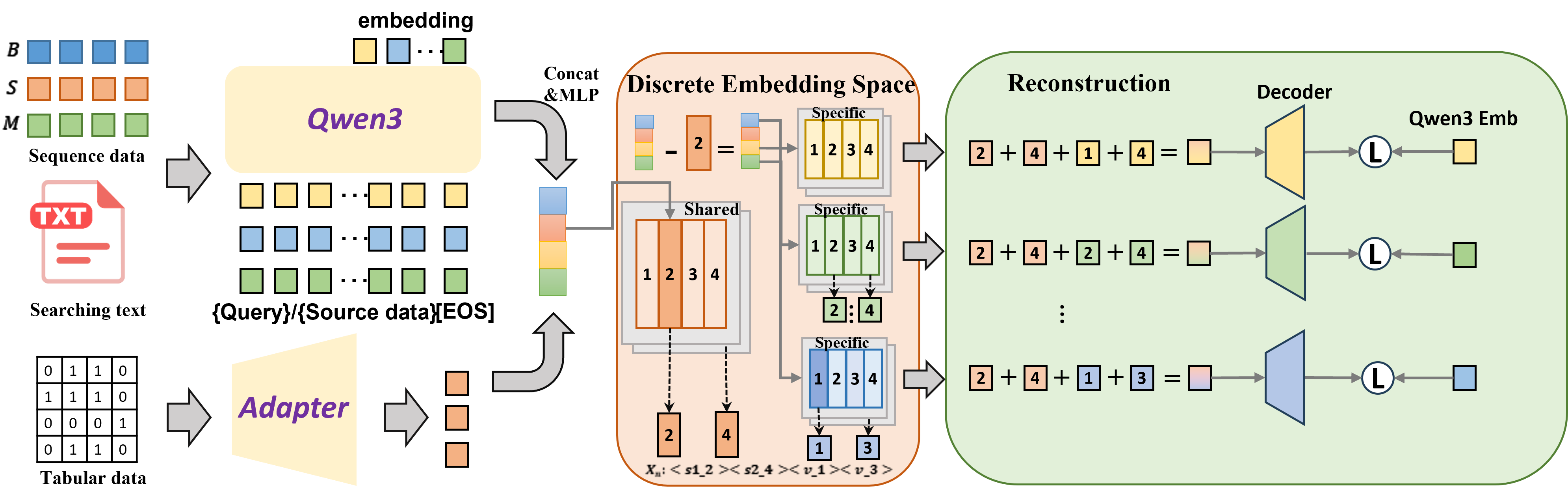}
    \caption{Overview of Unified User Tokenizer. First, we utilize Qwen3 embedding model to encode the long context feature into compact yet expressive embeddings. Then, we propose a MRQ-VAE with a shared-specific codebook hierarchy to further compress the data into discrete embedding. Finally, we reconstruct Qwen3 embeddng of multi-source data through source-specific MLP decoder.}
    \label{fig:overview}
\end{figure*}
\section{Methodology}\label{method}
This section presents FOUNDv2, a unified user modeling framework centered on the Unified User Tokenizer (U\textsuperscript{2}QT). As illustrated in Figure~\ref{fig:overview}, U\textsuperscript{2}QT transforms heterogeneous multi-source data into a concise, quantitative language via a two-stage compression paradigm. The first stage project diverse behavioral and textual inputs into an expressive latent language space to ensure semantic richness (Section~\ref{method:encoding}). The second stage then compresses these representations into a discrete embedding space through vector quantization, optimizing both storage efficiency and information density (Section~\ref{method:tokenizer}). Finally, we pre-train the framework using hierarchical multi-scale alignment tasks to capture user dynamics across different granularities (Section~\ref{method:pretext task}).

\subsection{Multi-source Encoding}\label{method:encoding}
Given the heterogeneity and information disparity across data sources, prior methods typically processed each source independently and encoded it into a latent representation to streamline downstream procedures~\cite{huang2020tabtransformer, devlin2019bert}. While this design simplifies the pipeline, it leaves a semantic gap across sources that these models struggle to bridge. To address this, we leverage LLMs to map multi-source data into a shared language space, producing richer and better-aligned embeddings. We use the Qwen3 Embedding model~\cite{qwen3emb} to encode multi-source data. We construct a prompt that combines the query with the corresponding source data and feed it to the model. Specifically, we show an example of SPM data $S_{n}$ as illustrated in Appendix ~\ref{section:Prompt Template}. 

Formally, final embeddings of multi-source data are derived from the hidden state
of the last layer corresponding to [EOS] token. For tabular data, we employ a Multilayer Perceptron (MLP) to map raw attributes into a latent vector, as LLMs frequently exhibit suboptimal performance in interpreting structured tabular information:
\begin{equation}
\begin{aligned}
    &\hat{H}^{(V)}_{n}=\operatorname{LLM}(Query, V_n, [EOS]),\\
    &\hat{H}^{(T)}_{n}=\operatorname{MLP}(T_n),\\
    &\hat{H}^{(R)}_{n}=\operatorname{LLM}(Query, R_n, [EOS])
\end{aligned}
\end{equation}
Where $\hat{H}^{(X)}_{n}\in \mathbb{R}^{1\times d}$ denotes the final representation of the multi-source input, and $X\in\{V,T,R\}$ indexes the source. This representation compresses very long contextual information into a single embedding. 

\subsection{Quantized User Tokenizers}\label{method:tokenizer}
After extracting the effective information from various source data, our aim is to further compress the representation from continuous space to discrete space. Existing approaches require considerable storage and computational resources to maintain pretrained user representations. Consequently, there is a pressing need for effective data compression techniques that can significantly decrease storage and computation requirements without compromising the representational capacity of user embeddings. Motivated by the heterogeneous multi-source nature of user data, we propose a Multi-view RQ-VAE (MRQ-VAE) with a shared-specific codebook hierarchy to compress latent representations into a discrete embedding space while explicitly modeling the heterogeneous multi-source nature of user data.

For each source $x \in \{V, T, R\}$, the transformed source-specific user behavior is defined as $\bar{H}_{n}^{(x)} = \mathcal{M} \circ \mathcal{P} \left( \hat{H}_{n}^{(x)} \right) \in \mathbb{R}^{d_c}$, where $\hat{H}_{n}^{(x)}$ denotes the learnable query vectors. The transformation is performed by first applying a mean pooling function $\mathcal{P}$ for semantic summarization, followed by a shared multi-layer perceptron $\mathcal{M}$ for dimensionality alignment. Then, MRQ-VAE first encodes it into a latent representation $z_{n}^{(x)}$. The quantization then occurs in two hierarchical stages: (i) Shared Codebook Stage ($L_c$ levels): Captures cross-source semantic coherence using a shared codebook $\mathcal{C}^{l,(S)}=\{v_{k}^{l,(S)}\}_{k=1}^{K}$, and (ii) Source-specific Codebook Stage ($L_u$ levels): Refines source-unique patterns using source-specific codebooks $\mathcal{C}^{l,(x)}=\{v_{k}^{l,(x)}\}_{k=1}^{K}$ for each source $x \in \{V, T, R\}$.

To reduce notational clutter, we omit symbols denoting source information and whether the codebook is shared or specific. The residual quantization process at level $l$ for user $n$ is defined as:
\begin{equation}
c^l_n = \mathop{\arg\min}\limits_{k} \parallel r_n^l - v_{k}^{l} \parallel_{2}^{2},
\end{equation}
\begin{equation}
r_n^{l+1} = r_n^l - v_{c_n^l}^l,
\end{equation}
where $c_n^l$ is the selected code index, $v_{k}^{l}$ is a vector from the relevant codebook ($\mathcal{C}^{l,(S)}$ or $\mathcal{C}^{l,(x)}$), and $r_n^l$ represents the residual vector. The initial residual is set to $r_n^0 = z_{n}$. 

After processing through all $L = L_c + L_u$ levels, where the first $L_c$ levels are shared and the remaining $L_u$ levels are specific to the source $x$, the quantized representation is obtained by summing the selected codebook vectors:
\begin{equation}
\hat{z}_{n}^{(x)} = \sum_{l=1}^{L_c} v_{ c_n^l}^{l,(S)} + \sum_{l=L_c+1}^{L_c+L_u} v_{c_n^l}^{l,(x)}.
\end{equation}
This design hierarchically disentangles common knowledge (learned via shared codebooks) from source-unique patterns (learned via specific codebooks).

The quantized representation $\hat{z}_{n}^{(x)}$ is then fed into a source-specific MLP decoder to reconstruct the Qwen3 embedding $\hat{H}_n^{(x)}$. For notational simplicity, we also omit symbols indicating source information and whether the codebook is shared or specific. The loss function comprises:
\begin{equation}
\mathcal{L}_{\text{re}} = \parallel\hat{X}_n - X_n\parallel_{2}^{2},
\end{equation}
\begin{equation}
\mathcal{L}_{\text{rq}} = \sum_{l=1}^{L} \left( \parallel \operatorname{sg}[r^l_n] - v_{c^l_n}^{l} \parallel_{2}^{2} + \alpha \parallel r_n^l - \operatorname{sg}[v_{c_n^l}^{l}] \parallel_{2}^{2} \right),
\end{equation}
where $\operatorname{sg}[\cdot]$ denotes the stop-gradient operator and $\alpha$ is a loss coefficient. The total loss for training the user tokenizer is $\mathcal{L}_{\text{re}} + \mathcal{L}_{\text{rq}}$, ensuring both accurate reconstruction of raw content and minimization of the distance between residuals and their assigned codebook vectors.

Once the unified user tokenizer is trained, we can translate the multi-souce heterogeneous data into a unified quantitative language space. For example, if we assign a two-level shared codebooks and two-level specific codebooks for each domain, we can represent a user as $U_n=\{\langle s_n^1 \rangle,\langle s_n^2 \rangle,\langle t_n^1 \rangle,\langle t_n^2 \rangle,\langle s_n^1 \rangle,\langle s_n^2 \rangle,\\\langle r_n^1 \rangle,\langle r_n^2 \rangle,\langle s_n^1 \rangle,\langle s_n^2 \rangle,\langle v_n^1 \rangle,\langle v_n^2 \rangle\}$, where $\langle s_n^i \rangle$ means the token from $i$-th level shared codebook and $\langle x_n^i \rangle$ denotes the token from $i$-th level $x$-specific codebook. However, translating the user representation into a unified quantitative language may lead to a collision in which different users are assigned the same tokens. To ensure industrial-grade reliability, we incorporate a fallback mechanism to resolve the infrequent occurrences of token collisions (occurring in <0.5\% of cases). This strategy strictly preserves the original business-relevant ordering, thereby guaranteeing the model's seamless deployability in high-concurrency production environments:
\begin{equation}
    (e_1,e_2,\dots,e_n)=\operatorname{argsort}_{\text{engage}}(U_1,U_2,\dots,U_n),
\end{equation}
where $\operatorname{argsort}_{\text{engage}}(\cdot)$ is the sorting function for designating a unique special token to user-defined tokenizers that generate overlapping or identical tokens, ensuring distinct identification in downstream processing. $e_k\in \mathbb{R}^{d_{cb}}, k\in [1, \dots,n]$ is the representation of each special token. Thus, the final user tokenizer is composed of domain-specific and special tokens, as $U_n=\{\langle s_n^1 \rangle,\dots,\langle t_n^2 \rangle,\langle s_n^1 \rangle,\dots,\langle r_n^2 \rangle,\langle s_n^1 \rangle,\dots,\langle v_n^2 \rangle,\langle e_n^1 \rangle\}$. \textit{We illustrate the advantage of our proposed framework in Appendix~\ref{section: advantage}.}

\begin{figure}
    \centering
    \includegraphics[width=1.0\linewidth]{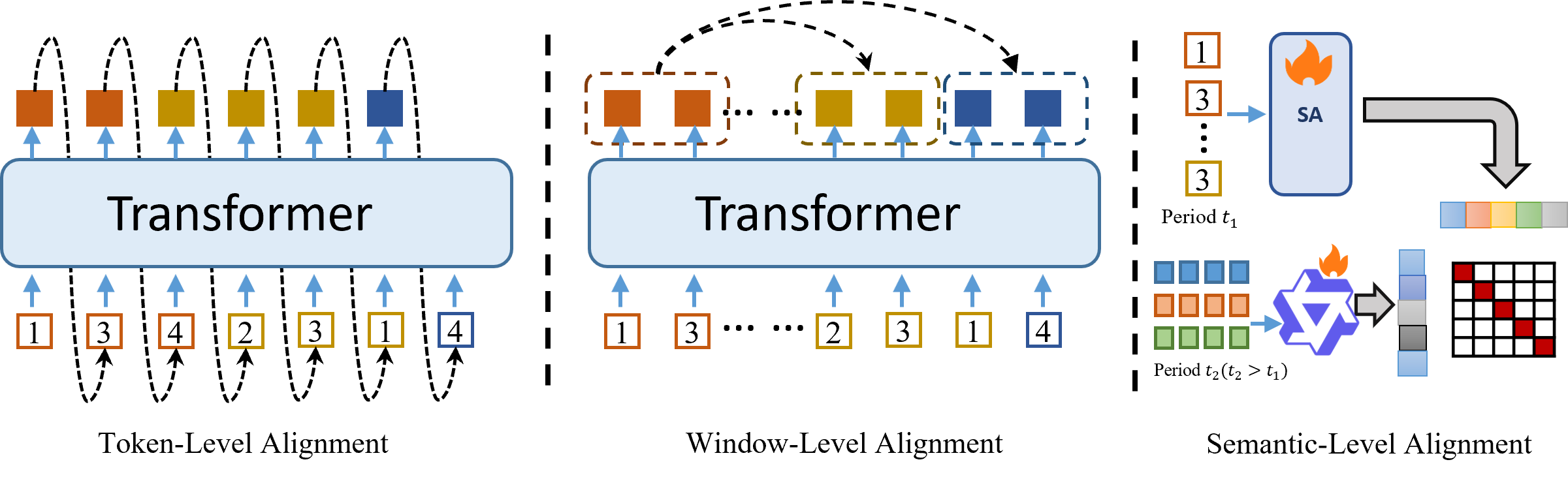}
    \caption{Multi-scale alignment task. We design token, window, and semantic-level alignment task to endow user tokenizer with fine-grained temporal intelligence.}
    \label{fig:pretext task}
\end{figure}

\subsection{Multi-Scale Alignment Pretext Task}\label{method:pretext task}
To characterize the complex dynamics of user behavior, we propose a suite of hierarchical alignment pretext tasks, spanning token-, window-, and semantic-level granularities. These tasks are designed to infuse the U\textsuperscript{2}QT tokenizer with fine-grained temporal intelligence and predictive capabilities. Specifically, token-level alignment is employed to capture local dependencies within individual behavioral events, while window-level alignment characterizes the long-term periodicity and cyclical rhythms inherent in user trajectories. Furthermore, we introduce a semantic-level alignment strategy, inspired by the cross-modal contrastive learning paradigm of CLIP, to ensure that the learned tokens remain predictive of future behavioral patterns within a high-level conceptual space.
\subsubsection{Token-Level Alignment}
The token-level alignment task is designed to capture latent structural dependencies within individual behavioral events. For a behavioral sequence decomposed into discrete tokens $\mathcal{Z} = \{z_1, z_2, \dots, z_K\}$, we implement an intra-behavioral coherence objective. Specifically, a self-supervised predictive loss is employed to ensure that tokens are semantically grounded in their local context:
\begin{equation}
    \mathcal{L}_{\text{token}} = \sum_{k=1}^{K-1} \omega_k (1 - \varphi(h_k, T_\theta(z_{k+1})))
\end{equation}
\begin{equation}
    \omega_k = 1 + k / \tau_t
\end{equation}
where $h_k$ denotes the representation for the $(k+1)$-th token predicted by the transformer, and $T_\theta(z_{k+1})$ represents its corresponding ground-truth embedding in pretrained codebook. We utilize the cosine similarity function $\varphi$ to quantify the alignment distance. Furthermore, a position-aware weight $\omega_k$ is introduced to assign higher importance to recent user tokens, with $\tau_t$ serving as a temperature hyperparameter.
\subsubsection{Window-Level Alignment}
To characterize macro-temporal patterns and behavioral periodicity (e.g., weekly or monthly routines), we design a window-level alignment task. We partition a user's long-term trajectory into consecutive temporal windows $\mathcal{W} = \{W_1, W_2, \dots, W_N\}$. To capture recurring behavioral motifs, we pull the representations of temporally correlated windows closer in the latent space:
\begin{equation}
    \mathcal{L}_{\text{win}} = -\frac{1}{B} \sum_{i=1}^{B} \log \frac{\exp(s(h_i^{W_a}, h_i^{W_b}) / \tau_w)}{\sum_{j=1}^{B} \exp(s(h_i^{W_a}, h_j^{W_b}) / \tau_w)}
\end{equation}
where $h_i^{W_a}$ and $h_i^{W_b}$ are the aggregated representations of two windows for user $i$ that exhibit periodic similarity (e.g., behaviors on the same day across different weeks). $s(\cdot, \cdot)$ denotes cosine similarity, and $\tau_w$ is a temperature hyperparameter. This allows U\textsuperscript{2}QT to internalize long-range periodicity beyond immediate actions.
\subsubsection{Semantic-Level Alignment}
Inspired by the cross-modal success of CLIP \cite{li2022blip, li2023blip2,zhu2025graphclip}, we align user representations with future behavioral descriptions in natural language to bridge the gap between discrete tokens and high-level intent. Specifically, we first transform future behavioral attributes at time $t_2$ into a structured narrative using a template function $\mathcal{F}$. These descriptions are then encoded into semantic embeddings $e_i^{nl}$ via a LoRA-tuned \cite{hu2022lora} LLM:
\begin{equation}
    e_i^{nl} = \operatorname{LLM}\left(\mathcal{F}(B_{i,t_2}, S_{i,t_2}, M_{i,t_2})\right)
\end{equation}
where $B, S, M$ represent purchased items, amount spent, and payment status, respectively. To align the fused tokenizer representation $e_{i,t_1}^f$ with the future semantic embedding $e_{i,t_2}^{nl}$, we employ the InfoNCE loss \cite{oord2019representationlearningcontrastivepredictive}:
\begin{equation}
    \mathcal{L}_{\text{sem}} = -\frac{1}{B} \sum_{i=1}^{B} \log \frac{\exp(s(e_{i,t_1}^f, e_{i,t_2}^{nl}) / \tau_s)}{\sum_{j=1}^{B} \exp(s(e_{i,t_1}^f, e_{j,t_2}^{nl}) / \tau_s)}
\end{equation}
The total pre-training objective of U\textsuperscript{2}QT is a joint optimization of the three levels:
\begin{equation}
    \mathcal{L}_{\text{total}} = \lambda_1 \mathcal{L}_{\text{token}} + \lambda_2 \mathcal{L}_{\text{win}} + \lambda_3 \mathcal{L}_{\text{sem}}
\end{equation}
where $\lambda_1, \lambda_2, \lambda_3$ are scaling coefficients to balance the alignment granularities.

\section{Experiment}
In this section, we conduct comprehensive experiments to answer following questions:
\begin{itemize}
    \item \textbf{RQ1}: How does our FOUNDv2 perform against various state-of-the-art user representation methods?
    \item \textbf{RQ2}: If our proposed module(e.g. multi-view RQ-VAE, multi-scale alignment tasks) works effectively?
    \item \textbf{RQ3}: What are the primary advantages of the compression mechanism achieved by Multi-view RQ-VAE?
    \item \textbf{RQ4}: What valuable insight can we achieve from well-learned codebook?
    \item \textbf{RQ5}: How does our proposed FOUNDv2 perform in industrial environment?
\end{itemize}
\begin{table}[]
    \caption{Data statistics for user tokenizer pretraining as well as the test benchmarks.}
\footnotesize
    \centering
    \begin{tabular}{ccccc}
    \toprule
         Dataset & No. & Domain & Scenario & Number \\
         \hline
         $\mathcal{D}_{train}$& - & General & General & $\approx$20 million \\
         \hline
         \multirow{5}{*}{$\mathcal{D}_{test}$} & 1 & future behavior & willing for takeout & $\approx$ 500,000\\
         & 2 & future behavior & purchasing power & $\approx$ 500,000\\
         & 3 & security &  money Laundering & $\approx$ 500,000\\
         & 4 & user profile & preference for games & $\approx$ 500,000\\
         \toprule
    \end{tabular}
    \label{tab:dataset}
\end{table}

\begin{table*}[]
    \caption{Quantitative comparison on downstream task performance on Alipay benchmarks. AUC / KS are presented.}
    \centering
    \begin{tabular}{ccccc}
    \toprule
         Methods & $\mathcal{D}_{test}$\#1 &  $\mathcal{D}_{test}$\#2 & $\mathcal{D}_{test}$\#3 & $\mathcal{D}_{test}$\#4 \\
         \hline
         One4all & 0.7663/0.3998 &  0.8348/0.4885 & 0.8855/0.6266 & 0.9750/0.8600 \\
         MSDP & 0.7891/0.4361 & 0.8464/0.4989 & 0.8810/0.6194 & 0.9731/0.8505 \\
         FOUND & 0.8315/0.5144 & 0.9381/0.7228 & 0.9026/0.6952 & 0.9425/0.8064 \\
         One4all+UT & 0.7752/0.4185 & 0.8649/0.5360 & 0.8961/0.6476 & 0.9658/0.8429 \\
         MSDP+UT & 0.7935/0.4427 & 0.8526/0.5087 & 0.8964/0.6479 & 0.9744/0.8588 \\
         FOUNDv2(RQ-VAE) & 0.8338/0.5174 & 0.9429/0.7313 & 0.9237/0.7079 & 0.9765/0.8631 \\
         \textbf{FOUNDv2(MRQ-VAE)} &  \textbf{0.8399/0.5262} & \textbf{0.9516/0.7585} & \textbf{0.9639/0.8342} & \textbf{0.9797/0.8747}\\
         \toprule
    \end{tabular}

    \label{tab:performance comparision}
\end{table*}

\subsection{Experimental Setup}
\subsubsection{Datasets}
We pretrain our framework on a large-scale Alipay dataset comprising approximately 20 million heterogeneous, multi-scenario user records ($\mathcal{D}_{train}$). 
The evaluation benchmark spans three primary domains: security risk control, behavior prediction, and user profiling. 
As detailed in Table~\ref{tab:dataset}, we select four representative tasks per domain to construct the test set $\mathcal{D}_{test}$, with each task containing roughly 500,000 records. 

\subsubsection{Baselines}\label{section:Baselines}
We compare our U\textsuperscript{2}QT with the following state-of-the-art methods in user representation.
\begin{itemize}
    \item \textbf{MSDP}~\cite{fu2023robust}: Multi-scale Stochastic Distribution Prediction model for learning user behavioral sequence representation, which takes the prediction on user’s behaviors distribution over a period of time as the self-supervision signal.
    \item \textbf{One4all}~\cite{shin2021one4all}:One4all is a general-purpose representation learning through large-scale pre-training, while U-MLP is one of its extended user targeting model which adds MLP user decoder to generate targeted users.
    \item \textbf{FOUND}~\cite{dou2025transferable}:FOUND is a general user modeling method pretrained by self-supervised stage and user-text alignment stage, which enables zero-shot transfer and few-shot user targeting.
\end{itemize}
We also utilize the pretrained user tokenizer in downstream recommendation task based on following model,
\begin{itemize}
    \item \textbf{MaskNet}~\cite{wang2021masknetintroducingfeaturewisemultiplication}: MaskNet introduces multiplicative operation into DNN ranking system by proposing instance-guided mask which performs element-wise product both on the feature embedding and feedforward layers.
\end{itemize}
\subsubsection{Implementation Detail}\label{section:Implementation Detail}
In this section, we present the detailed implementation of our FOUNDv2 framework. We first describe the hyperparameters used in our model. First, Qwen3-Embedding-0.6B\footnote{https://huggingface.co/Qwen/Qwen3-Embedding-0.6B} is utilized to derive embeddings in 1024-dim from mutli-source data. The Multi-view RQ-VAE module consists of a four-layer shared codebook with 1024 entries of 256 dimension, as well as six (one for each data source) 2-layer source-specific codebooks also comprising 1024 entries of 256 dimension. The hyperparameters $\lambda_1$, $\lambda_2$ and $\lambda_3$ are set to 0.2, 0.4 and 0.4, respectively. The model is optimized using AdamW with a learning rate of 1e-3. All experiments are conducted on a cluster of 16 PPU-810E processors, each with 100 GB of memory.

\subsubsection{Evaluation Metrics}\label{section:Evaluation Metrics}
For \textbf{User Representation} including future behavior and user profile prediction, linear probe representation analysis is conducted on all annotated datasets, which is evaluated by AUC (Area Under the ROC Curve~\cite{AUC}) and KS (Kolmogorov-Smirnov~\cite{KS}) metrics. For \textbf{Recommendation Task}, we utilize AUC and HR (Hit Rate) to evaluate the performance of models.

\subsection{Performance Analysis(RQ1)}
As illustrated in Table ~\ref{tab:performance comparision}, the experimental results demonstrate that FOUNDv2 significantly outperforms baseline methods across downstream tasks in our designed benchmark. Our model achieves superior performance in AUC and KS in all domains and outperforms the FOUND baseline 3\% / 9\% on average in AUC / KS. Moreover, we utilize the pretrained user tokenizer as supplementary feature in existing methods(e.g., One4all and MSDP) to verify the effectiveness. The result illustrates that these baselines gain huge improvement and outperform the original performance 1.2\%/3.0\% at AUC/KS for One4all and 0.8\%/2.3\% at AUC/KS for MSDP. Furthermore, FOUNDv2 integrated with our proposed MRQ-VAE consistently outperforms the variant utilizing vanilla RQ-VAE. This performance gain validates the efficacy of our hierarchical architecture, which leverages both shared and source-specific codebooks for mutli-source heterogeneous user data.

\subsection{Ablation Study(RQ2)}
In this section, all reported avg results are the average value over around 50 real-world application scenarios.

\subsubsection{Ablation study on codebook design}
We present an ablation study to evaluate the effect of codebook architecture. Specifically, we compare shared-only RQ-VAE and specific-only RQ-VAE with our proposed Multi-view RQ-VAE and we use 4 tokens to represent single user behavior. Additionally, we examine how performance scales with codebook capacity (number of layers, entries per layer, and code dimension). 

As shown in Table~\ref{tab:framework ablation}, the results yield two observations:

i) Our Multi-view RQ-VAE consistently outperforms the shared-only and specific-only RQ-VAE across all benchmarks. In the first two rows of Table~\ref{tab:framework ablation} (which hold the configuration fixed at 256 entries per layer and 128-dimensional code vectors), the proposed architecture achieves superior results, validating the effectiveness of its multi-view design.
ii) Performance scales positively with increased codebook capacity. Comparing the 256x128 configurations, moving from "2share+\\2specific" to "4share+2specific" yields gains on both metrics of 0.0015 AUC / 0.0037 KS, indicating that greater depth helps capture more complex patterns. Furthermore, increasing entries per layer from 256 to 1024 and the code dimension from 128 to 256 is also attributed to the performance improvement. For example, moving from "4share+2specific" (256x128) to "4share+2specific" (1024x256) produces substantial additional improvements of 0.0069 AUC and 0.0128 KS.


\subsubsection{Ablation study on multi-scale alignment tasks}
\begin{table}
    \caption{Ablation study on codebook design and scaling law of codebook volume.}
    \centering
    \begin{tabular}{cccc}
    \toprule
        Design & Volume & avg.AUC & avg.KS \\ \hline
        4share & 256$\times$128 & 0.7578 & 0.4213 \\ 
        4specific & 256$\times$128 & 0.7606  & 0.4251 \\ 
        2share+2specific & 256$\times$128 & 0.7662  & 0.4373 \\ 
        4share+2specific & 256$\times$128 & 0.7677 & 0.4410 \\ 
        \textbf{4share+2specific} & \textbf{1024$\times$256} & \textbf{0.7746} & \textbf{0.4538} \\ \toprule
    \end{tabular}
    \label{tab:framework ablation}
\end{table}
\begin{table*}
    \caption{Ablation study on heterogeneous multi-source data on Alipay benchmarks. (-) means deleting the corresponding data source. AUC / KS are presented.}
    \centering
    \begin{tabular}{ccccc}
    \toprule
         Data & $\mathcal{D}_{test}$\#1 &  $\mathcal{D}_{test}$\#2 & $\mathcal{D}_{test}$\#3 & $\mathcal{D}_{test}$\#4 \\
         \hline
          \textbf{All Sources} &  \textbf{0.8399/0.5262} & \textbf{0.9516/0.7585} & \textbf{0.9639/0.8342} & \textbf{0.9797/0.8747}\\
         (-)Tabular & 0.8249/0.4988 & 0.9191/0.6756 & 0.9524/0.8045 & 0.9780/0.8675 \\
         (-)App & 0.8226/0.4953 & 0.9081/0.6521 & 0.9500/0.7991 & 0.9779/0.8674 \\
         (-)MiniProgram & 0.8193/0.4907 & 0.8976/0.6313 & 0.9492/0.7966 & 0.9635/0.8190 \\
         (-)Search & 0.8172/0.4872 & 0.8917/0.6184 & 0.9474/0.7935 & 0.9629/0.8136 \\
         (-)Bill & 0.6200/0.1697 & 0.7883/0.4316 & 0.9314/0.7701 & 0.9619/0.8135 \\
         \toprule
    \end{tabular}
    \label{tab:source ablation}
\end{table*}
We evaluate the individual contributions of our pretext tasks in Table~\ref{tab:pretext}, leading to two primary observations. First, the integration of token-level and window-level alignment consistently yields performance gains across the benchmark. This confirms that capturing intra-behavioral dependencies and macro-temporal periodicity provides complementary insights that enhance the overall representation. Second, within the semantic-level task, we observe a clear scaling law with respect to sequence length: as the input trajectory extends, performance improves accordingly. This validates the framework’s ability to effectively ingest and distill richer user information from long-term historical data.

\begin{table}[]
    \caption{Ablation study on alignment pretext tasks. \textbf{S}, \textbf{T} and \textbf{W} represent Semantic, Token, and Window-level alignment tasks.}
    \centering
    \begin{tabular}{cccc}
    \toprule
         Task & Token Number & avg.AUC & avg.KS \\
         \hline
         -- & $\approx$1100 & 0.6985 & 0.3185 \\
         S & $\approx$1100 & 0.7679 &  0.4406 \\
         S+W& $\approx$1100 & 0.7707 & 0.4445 \\
         S+T & $\approx$1100 & 0.7708 & 0.4449 \\
         S+T+W & $\approx$1100 & 0.7732 & 0.4489 \\
         \textbf{S+T+W}& \textbf{$\approx$2200} & \textbf{0.7746} & \textbf{0.4538} \\
         \toprule
    \end{tabular}
    \label{tab:pretext}
\end{table}

\subsubsection{Ablation study on data source}
In this subsection, we conduct an ablation study to evaluate the contribution of each heterogeneous data source. The experiment involves systematically removing one data domain at a time and then pre-training a corresponding tokenizer for evaluation on our benchmarks. The results, summarized in Table ~\ref{tab:source ablation}, lead to the following key conclusions:

i) All data domains contribute positively to the tokenizer's overall effectiveness. As shown in Table \ref{tab:source ablation}, removing any single data source results in a performance degradation across all benchmark tasks. This underscores the synergistic value of integrating diverse data.

ii) Domain-specific data is particularly crucial for its corresponding downstream tasks. A salient example is the removal of the "Bill" data, which led to a significant performance drop on the "Willingness for Takeout" and "Purchasing Power" prediction tasks. This is likely because the bill data contains rich, explicit signals about purchasing behaviors, such as takeout orders and other item expenditures, which are directly relevant to these specific predictions. \textit{We also provide ablation study on various open-source encoder in Appendix~\ref{section:encoder}.}



\begin{table}[]
    \caption{Storage and computation analysis through comparison between FOUNDv2 and FOUND.}
    \centering
    \begin{tabular}{ccccc}
    \toprule
         Model & Data Volume & Training Time & Storage \\
         \hline
         FOUND& 20 million samples & about 7 hours &  240GB \\
         FOUNDv2 & 20 million samples & about 2 hours & 8.2GB \\
         \toprule
    \end{tabular}
    \label{tab:storage}
\end{table}
\begin{table}[]
    \caption{Information density analysis through comparison between FOUNDv2 and FOUND.}
    \centering
    \begin{tabular}{ccccc}
    \toprule
         Model & Tem. Span & GPU Mem.& avg.AUC & avg.KS\\
         \hline
         FOUND& 60 days & $\approx$60GB &  0.7613 & 0.4259\\
         FOUNDv2& 60 days & $\approx$3GB & 0.7633 & 0.4291\\
         FOUNDv2 & 90 days & $\approx$5GB & 0.7732 & 0.4489 \\
         \textbf{FOUNDv2} & \textbf{180 days} & \textbf{$\approx$10GB} & \textbf{0.7746} & \textbf{0.4538} \\
         \toprule
    \end{tabular}
    \label{tab:density}
\end{table}

\subsection{Efficiency Analysis(RQ3)}
\subsubsection{Storage and training efficiency}
As shown in Table~\ref{tab:storage}, our evaluation is based on behavioral logs from 20 million users collected over a 90-day period ($\approx$240 GB of raw data). While the original FOUND framework uses the full raw sequence to represent each user, FOUNDv2 compresses this into a compact tokenized form. This redesign yields dramatic gains: a 3.5× faster training time (2h vs. 7h) and a 30× smaller representation footprint (8.2 GB vs. 240 GB). These improvements demonstrate FOUNDv2’s strong scalability and efficiency.

\subsubsection{Information density}
Table~\ref{tab:density} validates the \textbf{densing principles} in Section~\ref{section:intro}. With a fixed batch size of 256, FOUNDv2 extends the inference temporal span from 60 days (FOUND) to 180 days, while keeping GPU memory modest ( $ \approx $ 10GB). This leap in \textbf{information density}, enabled by MRQ-VAE’s compact but powerful tokenization of sparse behaviors, yields measurable gains: AUC rises from 0.7732 to 0.7746 and KS from 0.4489 to 0.4538 when scaling from 90 to 180 days. Crucially, this is achieved without prohibitive overhead, breaking the sequence-length–memory trade-off. FOUNDv2 thus enables scalable, high-fidelity modeling of long-range user behavior in real-world e-commerce.

\section{Codebook Analysis(RQ4)}\label{section:codebook analysis}

\subsection{CodeBook Utilization}
\begin{figure}
    \centering
    \includegraphics[width=0.9\linewidth]{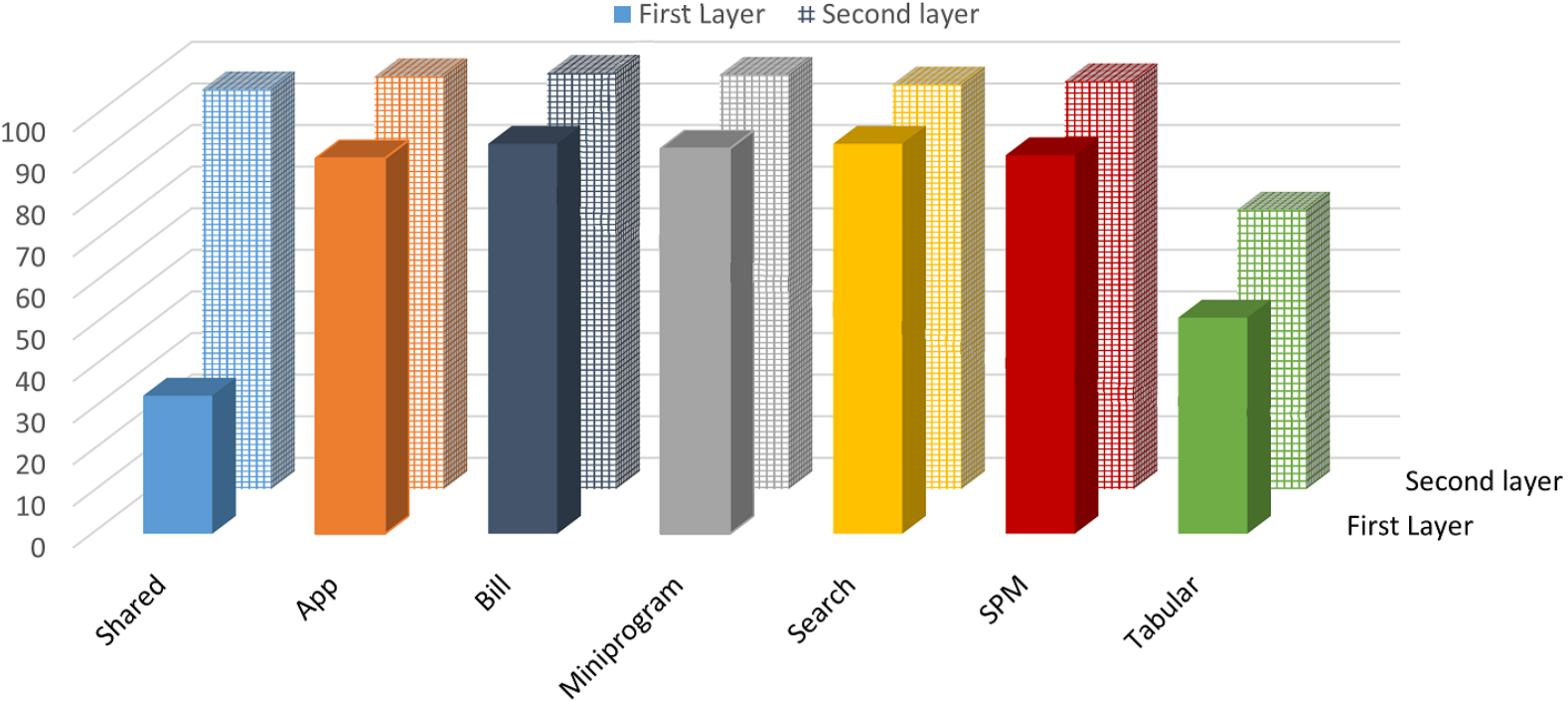}
    \caption{The utilization of each specific codebook at different layers.}
    \label{fig:codebook utilization}
    \vspace{-0.1cm}
\end{figure}

In this section, we analyze codebook utilization across quantization layers to understand model behavior. We define utilization as the percentage of unique codes used during training, indicating how effectively the model leverages its capacity~\cite{t-sne}. Results are visualized in Figure~\ref{fig:codebook utilization}. We draw two key observations:

i) \textbf{Hierarchical Abstraction}. Low shared codebook utilization in Layer 1 indicates a compact set of cross-domain patterns, while higher utilization in Layer 2 and domain-specific codebooks reflects refinement into diverse, fine-grained representations. This synergy enables efficient modeling of both universal and nuanced information.
    
ii) \textbf{Domain-dependent Capacity}. Utilization varies with data properties. The "Tabular" domain exhibits the lowest utilization, as numerical data has less semantic diversity than textual data (e.g., "App" or "Bill"). Consequently, it is well-represented by a smaller set of codes, aligning with observed patterns.
\begin{figure}
    \centering
    \subfigure[4 shared]{\includegraphics[scale=0.21]{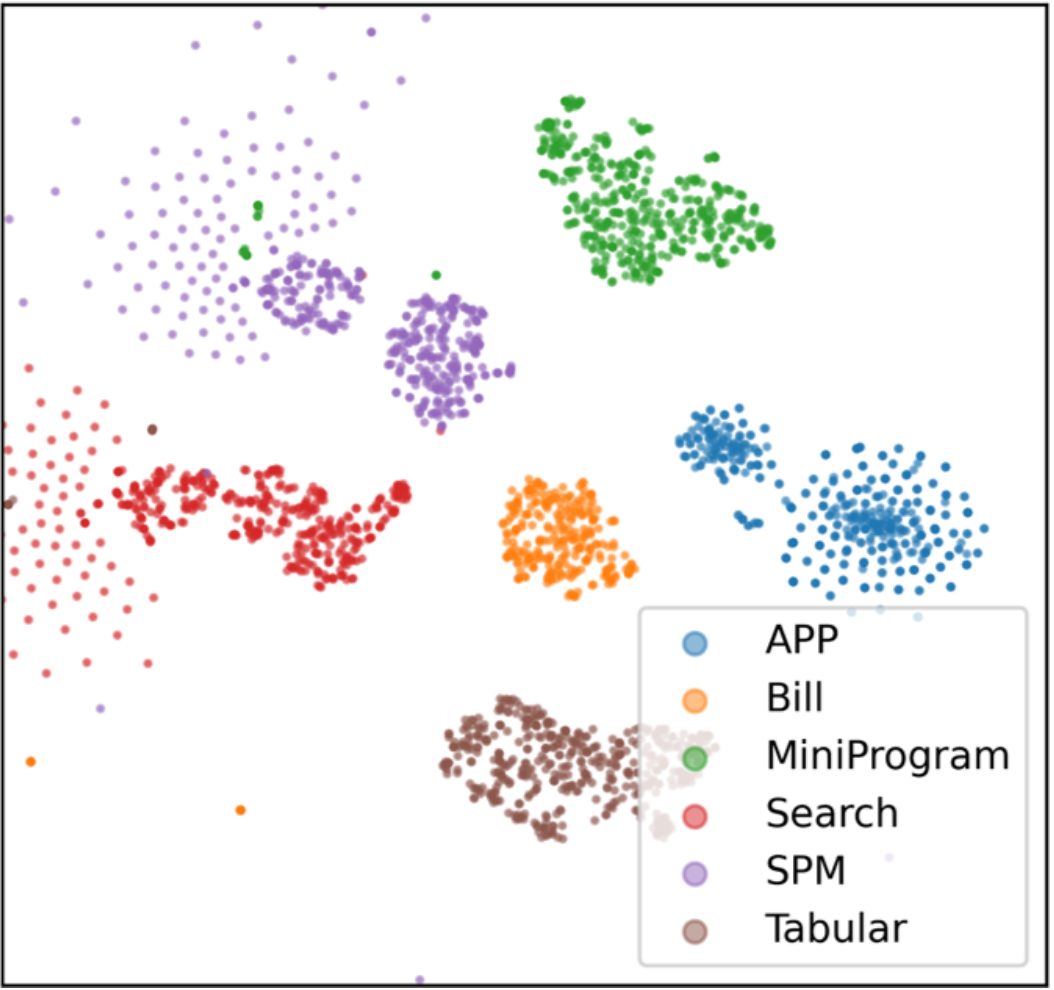}}
    \subfigure[2 shared + 2 specific]{\includegraphics[scale=0.21]{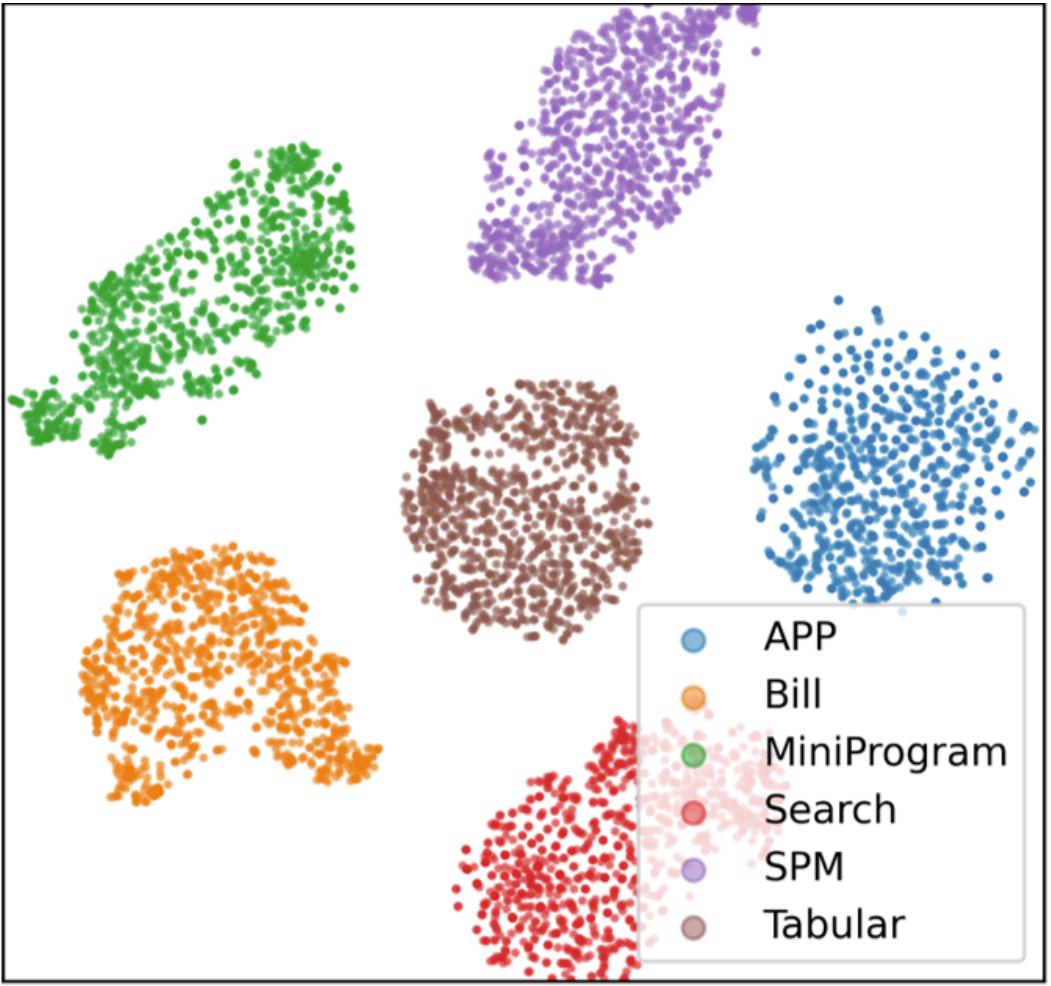}}
    \caption{Visualization of source-specific representation in different codebook design.}
    \label{fig:T-SNE}
    \vspace{-0.1cm}

\end{figure}

\subsection{Multi-source Visualization}\label{section:visualization}
Figure ~\ref{fig:T-SNE} presents a t-SNE visualization comparing the domain embeddings learned by our U\textsuperscript{2}QT model (b), which employs a 2 shared + 2 specific codebook architecture, against a baseline variant using only four shared codebook layers (a). The comparison reveals a stark contrast. In Figure ~\ref{fig:T-SNE}(b), our proposed architecture produces distinct, well-separated clusters for each domain. Each cluster exhibits high intra-domain cohesion (tightly packed points) and clear inter-domain separability (significant distance between clusters). Conversely, the purely shared architecture shown in Figure ~\ref{fig:T-SNE}(a) struggles to achieve this separation. The resulting clusters are visibly more dispersed and less defined, with ambiguous boundaries between several domains (e.g., Search and SPM). This qualitative evidence strongly validates our shared+specific design, proving its necessity for learning high-quality representations from heterogeneous data sources.

\section{Industrial Online A/B Testing(RQ5)}
To evaluate the practical effectiveness of \textbf{FOUNDv2} in production, we conducted extensive A/B tests across multiple high-impact industrial domains within the Alipay ecosystem. In the two primary transaction scenarios, \textbf{Scan-to-Pay} and \textbf{Tap-to-Pay}, we integrated the pretrained user representations from FOUNDv2 as core features into the PEPNet framework \cite{pepnet}. This deployment yielded a \textbf{15\% reduction} in coupon costs for Tap-to-Pay and a \textbf{4.71\% decrease} in pricing costs for Scan-to-Pay, demonstrating the model's ability to optimize resource allocation. 

In the \textbf{advertising} domain, we enhanced the recall strategy by replacing the representations from the state-of-the-art OmniRec model \cite{ominirec} with those pretrained by FOUNDv2. During the online A/B test conducted on the Alipay advertising platform, \textbf{FOUNDv2} demonstrated consistent performance gains across all primary monetization metrics. Specifically, the model achieved a \textbf{0.56\% improvement in Ad CPM} alongside a \textbf{0.50\% increase in ad consumption}. Furthermore, the advertisement-driven \textbf{transaction amount} rose by \textbf{0.72\%}, while the overall \textbf{customer value} exhibited a substantial \textbf{3.13\% growth}. These results collectively validate the scalability and superior representational power of FOUNDv2 in driving real world industrial value.

\section{Conclusion}
In this paper, we propose FOUNDv2, a comprehensive framework for user representation, centered on U\textsuperscript{2}QT(Unified User Quantized Tokenizer). To unify the representation of heterogeneous user behaviors from various sources, we utilize Qwen3 embedding model to encode them in language space. For more efficient storage and computation, we design a multi-view RQ-VAE that compresses user representations into discrete tokens further enhanced by multi-scale alignment tasks. We conduct comprehensive experiments across various downstream tasks to evaluate the effectiveness of our approach. The results consistently demonstrate that FOUNDv2 delivers strong representation capability, efficient storage and computation, and flexible cross-task generalization.

\begin{acks}
This work was supported by the Ant Group through Ant Research Intern Program.
\end{acks}

\clearpage

\bibliographystyle{ACM-Reference-Format}
\balance
\bibliography{reference}
\appendix
\section{Prompt Template}\label{section:Prompt Template}
\begin{promptbox}{Prompt Template}\label{method:template}
\textbf{Query}: \begin{CJK}{UTF8}{gbsn}
提取支付宝用户数据的文本特征
\end{CJK}

\textbf{Source Data}: \begin{CJK}{UTF8}{gbsn}
[dt][dt]付款码付款码付款码，2次[*]付款码付款码收钱码，1次[*]支付宝首页首页四大金刚，2次[*][dt][dt]]移动支付支付后推荐组件化模板内容组件单选组件，1次[*]移动支付支付后推荐组件化模板顶部区块，1次[*]2[dt][dt]小程序框架及组件授权页面获取基础能力授权，3次[*]实名认证开放认证授权页面开始认证，2次[*]支付宝医疗健康我的医保城市选择，1次[*]支付宝医疗健康我的页用户协议弹框，1次[*][dt][dt]
\end{CJK}
\end{promptbox}

\section{Ablation study on open-source encoder}\label{section:encoder}
\begin{figure}[!ht]
    \centering
    \includegraphics[width=1.0\linewidth]{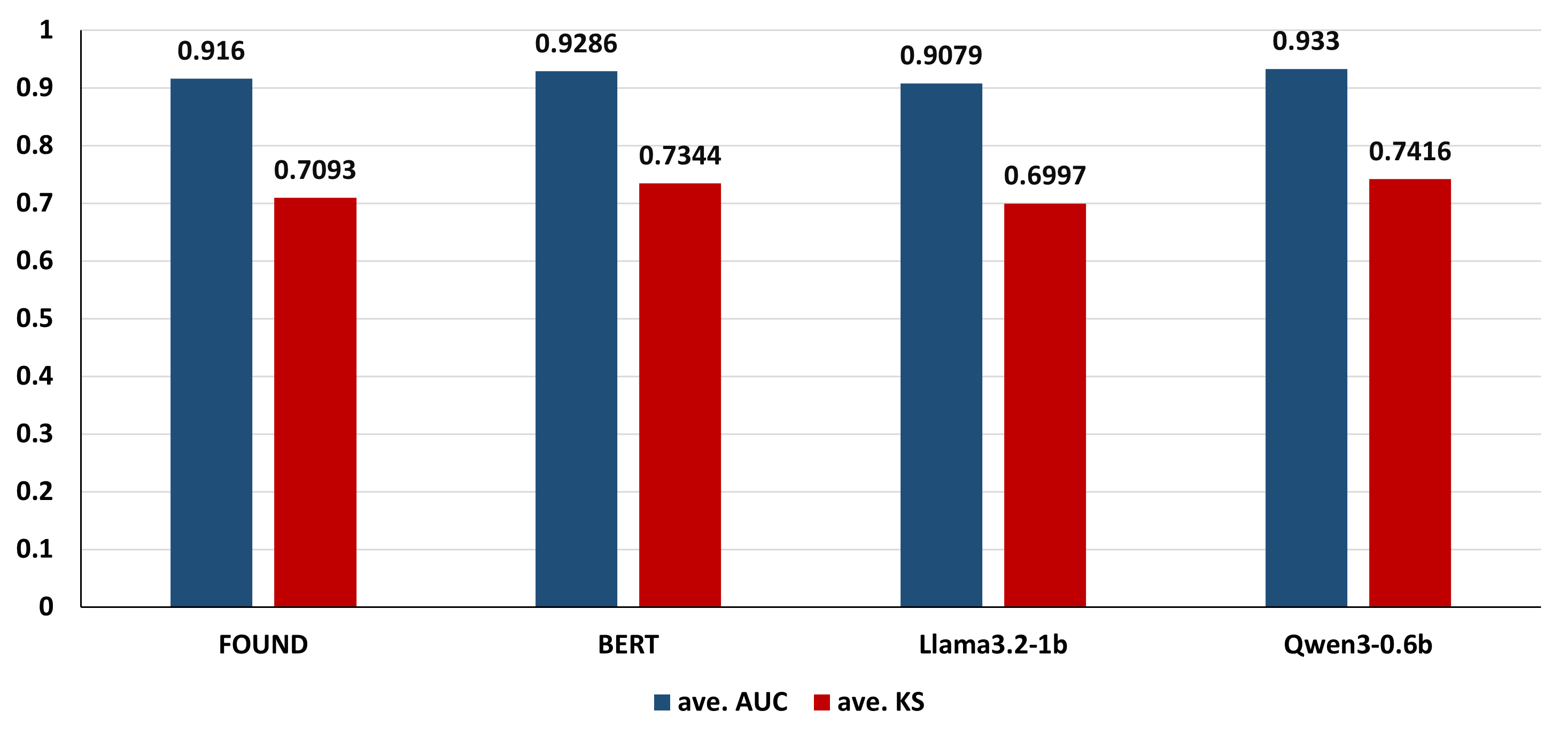}
    \caption{Ablation study on various open-source encoder.}
    \label{fig:ablation_encoder}
\end{figure}

To demonstrate the generalizability of U\textsuperscript{2}QT, we conduct experiments using various open-source encoders. As shown in the Figure~\ref{fig:ablation_encoder}, U\textsuperscript{2}QT exhibits remarkable robustness: even with a standard BERT backbone, it surpasses the FOUND baseline by 1.26\% in AUC. The performance drop observed with Llama 3.2-1B is likely due to its suboptimal language modeling capabilities in Chinese-specific contexts. However, the superior performance of the Qwen3-0.6B variant confirms that a high-quality, domain-compatible encoder can further amplify the benefits of our proposed quantization framework.

\section{Performance On Downstream Recommendation Task}
\subsection{Datasets}
For the downstream recommendation task, we adopt a chronological partition: interactions from the first 21 days (approximately 1{,}550{,}000 samples) form the training set, while the subsequent 3 days (approximately 100{,}000 samples) are split evenly into validation and test sets using stratified random sampling. This temporal split preserves interaction order to prevent data leakage and enhances robustness through dedicated holdout periods.

\subsection{Downstream Recommendation Task}
\begin{figure}[!ht]
    \centering
    \includegraphics[width=\linewidth]{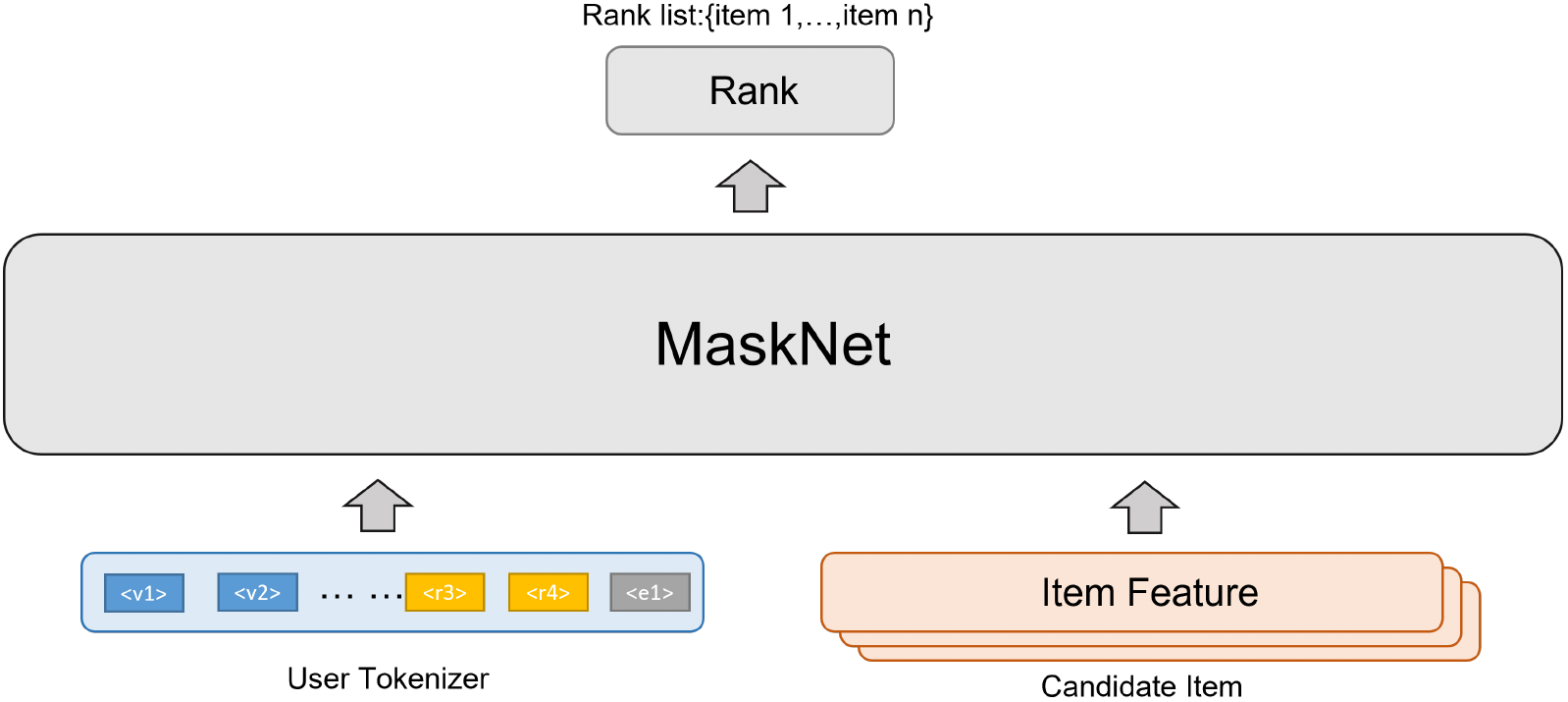}
    \caption{Application in recommendation task. We utilize the pretrained user tokenizer as user side feature in downstream recommendation task to verify the generalization of proposed method.}
    \label{fig:recommendation}
\end{figure}

In this section, we illustrate the usage of user tokenizer in downstream recommendation task. To describe the user-side feature, we employ the unified user tokenizer as a part of model input to representation multifaceted user information extracted from heterogeneous multi-source data as depicted in Figure ~\ref{fig:recommendation}. The overall process is as follows:
\begin{equation}
\begin{aligned}
    \text{UserRep}_i=[Code_1,\dots,Code_n]
\end{aligned}
\end{equation}
\begin{equation}
    \text{ItemRep}_j=\operatorname{Feat}(item_j)
\end{equation}
\begin{equation}
    y_{user_i,item_j}^{\text{CTR}}=\operatorname{MaskNet}(\text{UserRep}_i,\text{ItemRep}_j)
\end{equation}
where $\text{UserRep}_i$ denotes the user tokenizer of $U_i$ extracted from cashe storage, $\text{ItemRep}_j$ denotes the feature representation of $item_j$ and $\operatorname{MaskNet}(\cdot)$~\cite{wang2021masknetintroducingfeaturewisemultiplication} is a classical method in CTR prediction

\subsection{Performance Analysis}
We utilize our pretrained user tokenizer as user side feature in downstream recommendation task to verify the generalization of our proposed method. The result is illustrated in Table~\ref{tab:rectask}, showing that using tokenizer as a complementary feature achieves better performance and outperforms the baseline 0.7\% / 16.0\% / 65.5\% at AUC / HR@10 / HR@20.

\begin{table}[!ht]
    \caption{Quantitative comparison on recommendation task performance on Alipay benchmarks.}
    \centering
    \begin{tabular}{ccccc}
    \hline
         Feature & Backbone & AUC & HR@10 & HR@20 \\
         \hline
         User Profile& Masknet & 0.6476 & 0.2308 &  0.2968\\
         (+)Tokenizer & Masknet & 0.6527 & 0.2678 & 0.4913 \\
         \hline
    \end{tabular}
    \label{tab:rectask}
\end{table}

\section{Discussion on Architectural Synergy}\label{section: advantage}
The superiority of the proposed shared-plus-specific architecture over shared-only or specific-only configurations can be analyzed through the lenses of cross-source interaction and modality disentanglement. 

First, while a \textbf{shared-only} architecture facilitates global interactions by projecting all data sources into a single latent space, it inherently lacks the mechanisms required for modality disentanglement. In such a design, the common semantic space often becomes cluttered with source-specific noise, making it difficult for the model to isolate and preserve the unique structural nuances of individual behavioral domains. 

Second, a \textbf{specific-only} approach functions as a discrete counterpart to traditional multi-encoder paradigms where separate features are simply concatenated. Although this preserves the independence of each data source, it fails to capture the underlying semantic commonalities and synergistic relationships across different modalities. Without a shared component, the model cannot leverage the cross-source scaling law to build a truly universal representation space. 

In contrast, our \textbf{shared+specific} design strikes an optimal balance by integrating both philosophies. The shared codebooks extract high-level semantic universals that transcend individual sources, while the source-specific codebooks ensure that unique behavioral patterns are effectively disentangled and preserved. This hybrid structure enables FOUNDv2 to benefit from rich cross-source interactions without compromising the integrity of modality-specific information, ultimately leading to more robust and expressive user representations.




\section{Deployment Detail}\label{section:deployment}
To maintain up-to-date user representations across our massive user base of 1 billion, \textbf{FOUNDv2} adopts an efficient incremental update strategy in the production environment. 

On a daily basis, the system records multi-modal user behaviors and processes only the current day's activities. For each modality, the encoder generates 4 discrete tokens. These newly generated tokens are then concatenated to the existing historical token sequence, effectively updating the user's profile without re-processing past behaviors. 

This incremental deployment offers a significant efficiency advantage over the original \textbf{FOUND} framework. While the previous version required full inference across the entire user base of 1 billion to refresh representations, \textbf{FOUNDv2} restricts the computational workload to daily active users (approximately 400--500 million). By avoiding redundant calculations of historical data, we achieve a substantial reduction in daily inference latency and resource consumption, ensuring high-throughput serving for downstream large-scale industrial tasks.

\end{document}